\documentclass[letterpaper, 10 pt, conference]{ieeeconf}  % Comment this line out if you need a4paper

\IEEEoverridecommandlockouts                              % This command is only needed if 
\pdfoutput=1
\usepackage{amssymb}
\usepackage[english]{babel}
\usepackage[utf8]{inputenc}
\usepackage{amsmath}
\usepackage{graphicx}
\usepackage[colorinlistoftodos]{todonotes}
\usepackage{hyperref}
\usepackage{algorithm}
\usepackage{stfloats}
\usepackage{cooltooltips}
\usepackage{booktabs}
\usepackage{comment}

\title{Adapting Low-Cost Platforms for Robotics Research}

\author{Karimpanal T.G., Chamambaz M., Li W.Z., Jeruzalski T., Gupta A., Wilhelm E.}
\newcommand{\manualkeywords}[1]{\textbf{\textit{Index Terms ---}} #1}
\date{\today}

\begin{document}
\maketitle

\begin{abstract}
Validation of robotics theory on real-world hardware platforms is important to prove the practical feasibility of algorithms.  This paper discusses some of the lessons learned while adapting the EvoBot, a low-cost robotics platform that we designed and prototyped, for research in diverse areas in robotics.  
The EvoBot platform was designed to be a low-cost, open source, general purpose robotics platform intended to enable testing and validation of algorithms from a wide variety of sub-fields of robotics. 
Throughout the paper, we outline and discuss some common failures, practical limitations and inconsistencies between theory and practice that one  may encounter while adapting such low-cost platforms for robotics research. We demonstrate these aspects through four representative common robotics tasks- localization, real-time control, swarm consensus and path planning applications, performed using the EvoBots. We also propose some potential solutions to the encountered problems and try to generalize them.\\
%including sensor fusion, real-time control, swarming and artificial intelligence. 
%This paper introduces the EvoBot platform which was designed to be a low-cost, open source, general purpose robotics platform intended to enable testing and validation of research results in a wide variety of sub-fields of robotics. The paper outlines many lessons learned while developing this platform, as well as the lessons learned while applying it for state estimation, real-time control, swarming, and artificial intelligence research applications. Important considerations when adapting arbitrary low-cost platforms for robotics research are discussed in the context of four representative tasks from different common areas of robotics.\\
\manualkeywords{\textbf{low cost, open-source, swarm robotics, lessons learned}}

\end{abstract}

\section{Introduction}
The Motion, Energy and Control (MEC) Lab at the Singapore University of Technology and Design (SUTD) is engaged in an ambitious research project to create a swarm of autonomous intelligence-gathering robots for indoor environments. After reviewing commercially available low-cost robotics platforms, some of which are shown in Table \ref{table:prev_bots}, the decision was taken to build a custom ground-based robot platform for performing swarming research, dubbed the 'EvoBot'. The primary trade-off is between platform flexibility and cost, where existing robots are intended to either be applied in specific research areas, and are hence equipped with limited and specific sensor and communication capabilities,  or are more flexible with respect to sensor and firmware packages, but are also more expensive. In terms of software control development environments, well-established frameworks such as ROS  \cite{ROS} or the Robotics Toolbox \cite{Corke11a} have extremely useful modular functions for performing baseline robotics tasks, but require substantial modification between applications and require specific operating systems and release versions. 

With this in mind, the MEC Lab set out to develop the EvoBot platform with the following goals:
\begin{itemize}

\item Low cost:  affordable for research groups requiring a large number of swarming robots (e.g. more than 50) with sufficient sensing and control features.

\item Open source: The hardware, body/chassis design, application software and firmware for the EvoBots are fully open source in order to enable any group to replicate the platform with minimal effort.
All the hardware and software files used for the design of the EvoBots is available here:\newline  \href{https://github.com/SUTDMEC/EvoBot_Downloads.git}{https://github.com/SUTDMEC/EvoBot\_Downloads.git}

\item Adaptable: The final platform is intended to be as general purpose as possible, with minimum changes needed to be made to the base firmware by users in order to scale to a wide variety of common research applications. Some representative applications are described in section \ref{applications}. \ldots

\end{itemize}

In the process of designing the EvoBots platform, a great deal of failure-based learning was involved, and this paper compiles and synthesizes this process in order for it to be useful for future robotics researchers who intend to develop their own platforms, or adapt commercially available platforms. Specific emphasis will be on challenges common to robotic platforms in general, and approaches used to avoid failure when attempting to solve them. The authors would like to point out that each robotics project presents a unique sets of problems, so we have attempted to only describe problems we think may be generalized.

\section{Precedents and EvoBot Design}

\begin{figure}[b]
  \centering
  \includegraphics[width=0.5\textwidth]{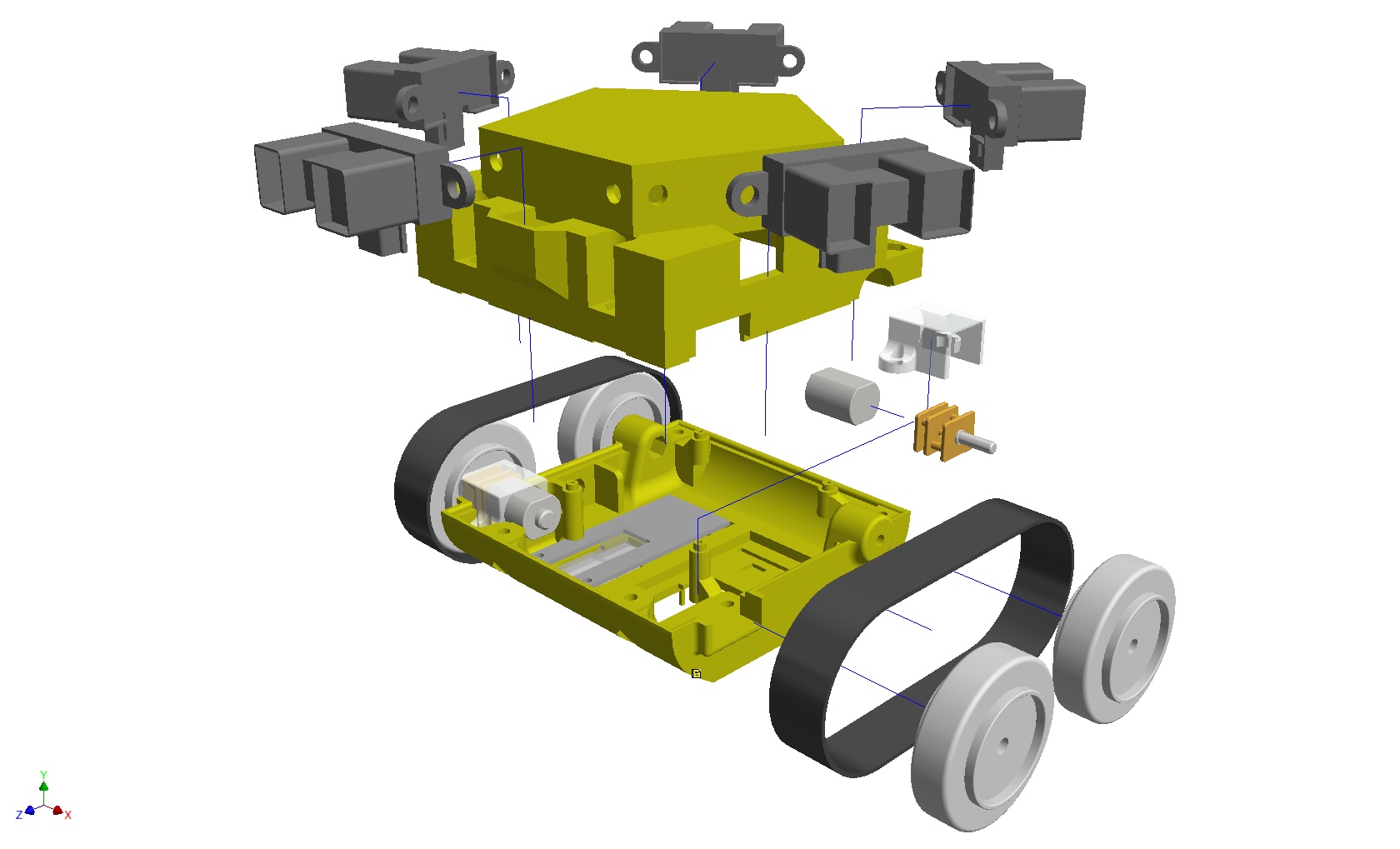}\\
  \caption{The 3-D printed case has two slots at the bottom for the optical flow sensors, a housing for the left and right tread encoders, and 5 IR depth sensors. The encoders on the forward wheels and the optional ultrasonic sensors are not shown}\label{fig:BotPic}
\end{figure}

This section will discuss commonalities and differences between the EvoBot and other comparable low-cost robotics platforms, as well as provide insights into the lessons learned during the design process.
In order to reduce the time between design cycles, the EvoBots were prototyped with a 3D printed body and developed across three major generations and several minor revisions. An exploded view of the EvoBot is shown in Figure \ref{fig:BotPic}.

%The EvoBot is a simple, low-cost platform intended to enable prototyping and validation in a variety of fields in robotics research such as control theory, adaptive autonomy and swarm robotics. Its compatibility with softwares such as $MATLAB$ makes it convenient for one to rapidly validate theoretical developments through experiments. In addition, the wireless communication capabilities of the EvoBots open up the possibility of using it as a prototyping platform for research in robotic swarm behaviour and control.\\

\begin{table}[h!]
\begin{scriptsize}
\begin{center}
\caption{Precedents for low-cost research robotics platforms}
\label{table:prev_bots}
\scalebox{0.9}{
    \begin{tabular}{cccccc}
    \toprule
	Platform & Est. Cost (USD) & Commercial & Scholar Hits &  Schematics & Code \tabularnewline
    \midrule
    \midrule
    Khepera \cite{mondada1999development} & 2200 & Yes & \textgreater 1000 & No & Yes\tabularnewline
    \midrule
	Kilobot \cite{rubenstein2012kilobot} & 100+ & Yes & 82 & No & Yes \tabularnewline
    \midrule
    e-Puck \cite{mondada2009puck} & 340+ & Yes & \textgreater 1000 & Yes & Yes \tabularnewline
    \midrule
    Jasmine \cite{arvin2011imitation} & 150+ & No & \textgreater 1000 & Yes & Yes \tabularnewline
    \midrule
    Formica \cite{english2008strategies} & 50 & No & 10 & Yes & Yes \tabularnewline
    \midrule
    Wolfbot \cite{betthauser2014wolfbot} & 550+ & No & 4 & Yes & Yes \tabularnewline
    \midrule
    Colias \cite{arvin2014development} & 50+ & No & \textgreater 1000 & No & No \tabularnewline
    \midrule
    Finch \cite{lauwers2010designing} & 100 & Yes & 73 & No & Yes \tabularnewline
    \midrule
    Amigobot \cite{adept_mobilerobots_amigobot_2014} & 2500 & Yes & 170 & No & Yes \tabularnewline
    \midrule
    EvoBot & 300 & Yes & 0 & Yes & Yes \tabularnewline
    \bottomrule
    \end{tabular}    
}
\end{center}
\end{scriptsize}
\end{table}

Like the Khepera, Finch, Amigobot (table \ref{table:prev_bots}) and most of the other platforms, locomotion on the EvoBot is achieved using a differential drive system, with motors on either side of its chassis. Although this system introduces kinematic constraints by restricting sideways motion, the associated simplicity in manufacturing and assembly, and in the mathematical model for use in control applications are significant advantages. The wheels are coupled to the motors through a gearbox with a gear reduction ratio of 1:100. After reduction, the final speed of the robot can be varied from -180 to 180 $mm/s$, so that both forward and backward motions are possible.\\ 
%This speed, while being similar to other platforms, failed to suffice for our applications. To mitigate this failure, a drop-in replacement motor and geartrain with a lower 1:50 reduction was used. GENERAL LEARNING 1: specify components with drop-in replacements \\ 
The speed of each motor is controlled by a pulse width modulated voltage signal. In order to ensure predictable motion, an internal PI controller is implemented by taking feedback from encoders that track the wheel movement. The controller parameters may need to be hand-tuned to compensate for minor mechanical differences between the two sides of the robot, arising from imperfect fabrication and assembly. \\
%To overcome this issue, calibration was performed using a special set of firmware, and a test jig where the devices were calibrated. GENERAL LEARNING 2: automate calibration process\\
%For obstacle detection and mapping applications, infra-red (IR) sensors were used. The 5 IR sensors, when calibrated, can sense in the range of 20-70 cm, leaving an undesirable 'dead zone' in sensing which must be compensated by the controller. The ultrasonic sensors, while not suffering from dead-zones, are prone to false readings from back-scatter sound energy. 
For obstacle detection and mapping applications, 5 infra-red (IR) sensors were placed on the sides of a regular pentagon to ensure maximum coverage, with one IR sensor facing the forward direction. Similar arrangement of range sensors is found in platforms such as Colias and e-puck (table \ref{table:prev_bots}). Despite this arrangement, there exist blind spots between adjacent sensors, which could lead to obstacles not being detected in certain orientations of the robot relative to an obstacle. The use of ultrasonic sensors instead of IRs, could reduce the probability of occurrence of blind spots due to their larger range and coverage, but they are also more expensive. It is thus advisable to choose sensors appropriately and to thoroughly test their performance and limitations before a final decision on the physical design and sensor placements is made.\\
Another feature present in the EvoBots common to other platforms is the use an inertial motion unit (IMU). The 6DOF IMU gives information regarding the acceleration of the robot in the x, y and z directions, along with roll, pitch and yaw (heading) information. Although in theory, the position of the robot can be inferred using the IMU data, in practice, these sensors are very noisy, and the errors in the position estimates from these sensors are unacceptably high even after using methods such as the Kalman filter \cite{julier_new_1995}. Although obtaining the position estimates from the IMU is not recommended in general, good estimates can be obtained with units that are capable of a much higher resolution and data rate. Such units however, come at an expense and thus may not be the ideal choice for a low-cost platform.

\subsection{Sensing Features}

While the sensors mentioned earlier in this section focused on features shared with other platforms, there are some sensing, control and communication features specific to the EvoBots.\\
In almost all robotics applications, errors in the robot's position estimates may gradually accumulate (e.g. for ground robots, skidding or slippage of the wheels on the ground surface confound the wheel encoders). In the first two generations of the EvoBots, this error in the wheel encoders caused substantial challenges for  localization.  In order to tackle this problem, optical flow sensors were placed on the underside of the EvoBot. They detect a change in the position of the robot by sensing the relative motion of the robot body with respect to the ground surface. This ensures that the localization can be performed more reliably even in the case of slippage.
As shown in Figure \ref{fig:BotPic}, there are two such sensors placed side by side at a specific distance apart. This arrangement allows inference of heading information of the robot along with information of the distance traveled.
Although the optical flow sensors perform well in a given environment, they need to be calibrated extensively through empirical means. In addition, it was found that the calibration procedure had to be repeated each time a new surface is encountered, as the associated parameters are likely to change from surface to surface.

As described above, the wheel encoders, optical flow as well as the IMU sensors can all be used to estimate the robot's position and heading. Each of these becomes relevant in certain specific situations. For example, when the robot has been lifted off the ground, the IMU provides the most reliable estimate of orientation; when the robot is on the ground and there is slippage, the optical flow estimates are the most reliable; and when there is no slippage, the position estimates from the encoders are reliable. Some details on the use of multiple sensors for localization are mentioned in section \ref{state_est}.

In their current configuration, the EvoBots also include the AI-ball, a miniature wifi-enabled video camera with comprehensive driver support and a low cost point to capture video and image data \cite{trek_sa_ai-ball_2014}. The camera unit is independent of the rest of the hardware and thus can be removed without any inconvenience if it is not needed.

\subsection{Control and Communication Features}

There are a wide variety of microcontrollers which are presently available on the market. The chipset used on the EvoBot is the Cortex M0 processor, which has a number of favorable characteristics such as low power, high performance and at the same time, is cost efficient, with a large number of available I/O (input/output) pins. One of the primary advantages of this processor is that it is compatible with the mbed platform (\href{https://developer.mbed.org/}{https://developer.mbed.org/}), which is a convenient, online software development platform for ARM Cortex-M microcontrollers. The mbed development platform has built-in libraries for the drivers, for the motor controller, Bluetooth module and the other sensors.

Like the e-puck, the EvoBots have Bluetooth communication capability. After experimenting with various peer-to-peer and mesh architectures, it was determined that in order to maintain flexibility with respect to a large variety of potential target applications, having the EvoBots communicate to a central server via standard Bluetooth in a star network topology using the low-cost HC-06 Bluetooth module was the best solution. Sensor data from the various sensors gathered during motion is transmitted to the central server every 70$ms$. The camera module operates in parallel and uses Wi-Fi 802.11b to transfer the video feed to the central server. The Bluetooth star network is also used to issue control commands to the robots. %This approach sacrifices communication range for communication through-put and timeliness. 
%The meshing protocols do not have the bandwidth to accomplish applications where mapping should be performed in real-time, for example.

In addition, having a star network topology with a central server indirectly allows additional flexibility in terms of programming languages. Only a Bluetooth link is required, and all the computation can be done on the central server. For example, the EvoBot can be controlled using several programming languages, including (but not limited to) $C$, $python$ and $MATLAB$. The Bluetooth link also opens up the possibility of developing smartphone mobile applications for it. Table \ref{table:sens_tab} summarizes all the sensors used in the EvoBot.

%The sensors used in the EvoBot are listed in table \ref{table:sens_tab}. Some of the sensors such as the optical flow (ADNS5090), wheel encoders and IMU (MPU6050) are used for localization, whereas some others such as the current sensor (ACS712) and voltage sensor are used for the power characterization of the robot. The proximity sensor (GP2Y0A02YK0F) is an infrared range sensor that is used to sense the distances to obstacles in the environment.

\begin{table}[h!]
\begin{scriptsize}
\caption{The final sensing capabilities of the 3rd generation EvoBot platform}
\label{table:sens_tab}
\begin{center}
\scalebox{0.95}{
    \begin{tabular}{cccc}
    \toprule
    Signal & Sensor & Frequency & Full Range Accuracy \tabularnewline
    \midrule
    \midrule
    acceleration & MPU6050 & 1000 Hz & +/- 0.1\% \tabularnewline
    \midrule
    Temperature & MPU6050 & max 40 Hz & +/-$1\,^{\circ}{\rm C}$ from -40 to $85\,^{\circ}{\rm C}$ \tabularnewline		
    \midrule
    x/y pixels & ADNS5090 & 1000 Hz& +/- 5mm/m \tabularnewline
    \midrule
    Encoders & GP2S60 & 400 Hz& +/- 5mm/m  \tabularnewline
    \midrule
    Battery current & ACS712 & 33 Hz& +/- 1.5 \% @ 25C  \tabularnewline
    \midrule
    Proximity & GP2Y0A02YK0F & 25 Hz& +/- 1mm  \tabularnewline
    \midrule
    Camera & AI-ball & 30fps & VGA 640x480   \tabularnewline
    \bottomrule
    \end{tabular}
}   
\end{center}
\end{scriptsize}
\end{table}

\section{General Problems} \label{general}

The EvoBot platform was designed to be useful for a wide range of research problems, and to enable researchers working in theoretical robotics to easily shift their algorithms from simulation to hardware. Moving from virtual to real platforms introduces a set of additional issues which must be considered. Some of these, which were encountered while developing applications for the EvoBots, are discussed in this section.

\subsection{Sensing}

While it may be considered acceptable to use ideal sensor models in simulation, in real-world robotics applications, and especially in the case of low-cost platforms, sensing information about the environment can be done only with a certain degree of range and accuracy.  It is thus important to be judicious in choosing the location of the sensors. For the case of range sensors for obstacle detection, blind spots must be avoided as much as possible to prevent unexpected behaviours. Also, the upper and lower limits of the sensor range must be considered. As in the case of the EvoBots or other low-cost platforms, the sensor range may be quite limited and prone to noise, and blind spots exist despite careful sensor placement. In case of occurrence of blind spots, it may be worth encoding default behaviours into the robot such as conditional wall following or automatic steering away from detected obstacles.

\subsection{Calibration} 

The behaviour of sensors are typically considered to be same for different environments during simulation. In practice, most sensors need some form of calibration. In addition, the calibration parameters for certain sensors may depend on the environment in which they operate. For example, the calibration parameters for the optical flow sensors in the EvoBots vary based on the surface on which they operate. In addition, the parameters are very sensitive to small variations in ground clearance which can be caused by variations in the robot body due to imperfect 3D printing or during assembly.  \\ It is often the case that parameters will have to be set and reset from time to time. This can considerably add to development time and effort. Automating the calibration process should be done whenever possible. If automation is not possible, one way to mitigate this issue is to have the software package include user commands that can change calibration parameters on the fly. This was the solution adopted for the EvoBots in order to calibrate the encoders and optical flow sensors. 

\subsection{Timing} 

In simulation, the robot's update loop and solvers can have either fixed or variable time steps, but it is most likely to be variable in practice. Thus, keeping track of time is critical, especially for real-time applications. For example, in the EvoBots, the performance of the localization algorithm (section \ref{state_est}) significantly depreciates if the time step is considered fixed.
%This problem is common to all platforms. 

\subsection{Communication} 

When a large amount of sensor data and instructions needs to be exchanged between platforms or between a platform and a central machine, maintaining the integrity of the communicated data is critical. When the exchanged information is delayed or lost either partially or completely during information exchange, it could lead to a series of unwanted situations such as improper formatting of data, lack of synchronization, incorrect data etc., These aspects are seldom accounted for in simulation. In the EvoBots, the issue of lack of data integrity was tackled by performing cyclic redundancy checks (CRC) \cite{4066263} as part of the communication routine. It is ideal to perform bit-wise checks as well, but this has not been done in the case of the EvoBots.

The issues listed in this section apply to almost all robotics applications in general. The following section discusses some of the applications and the associated issues encountered during their implementation using the EvoBots.

\section{Applications} \label{applications}
This section discusses application-specific issues in diverse sub-fields of robotics such as real-time control, swarm and artificial intelligence applications. These applications were selected because they are commonly implemented tasks from a wide variety of sub-fields in robotics.

\subsection{Localization} \label{state_est}
Ground robots typically need to have an estimate of position and heading. The EvoBot platform, as well as many other robots, are designed to be used in an indoor environment, and therefore it is not possible to use GPS without modifying the environment in question; also, positioning information provided by GPS has the accuracy in the order of a 10\textsuperscript{1} meters which is not suitable for small ground robots. For this reason, the set of on-board sensors are used to estimate the robot's states of position and heading. The on-board sensors however, are typically associated with some noise and retrieving positioning information from them leads to erroneous results. For instance, integrating acceleration data to get the velocity and position does not provide accurate state estimation and leads to huge offsets from the true value due to the associated noise. To overcome this issue, state estimation is performed using a method based on the Extended Kalman Filter \cite{julier_new_1995}.

\begin{figure}[!t]
  \centering
  \includegraphics[width=0.5\textwidth]{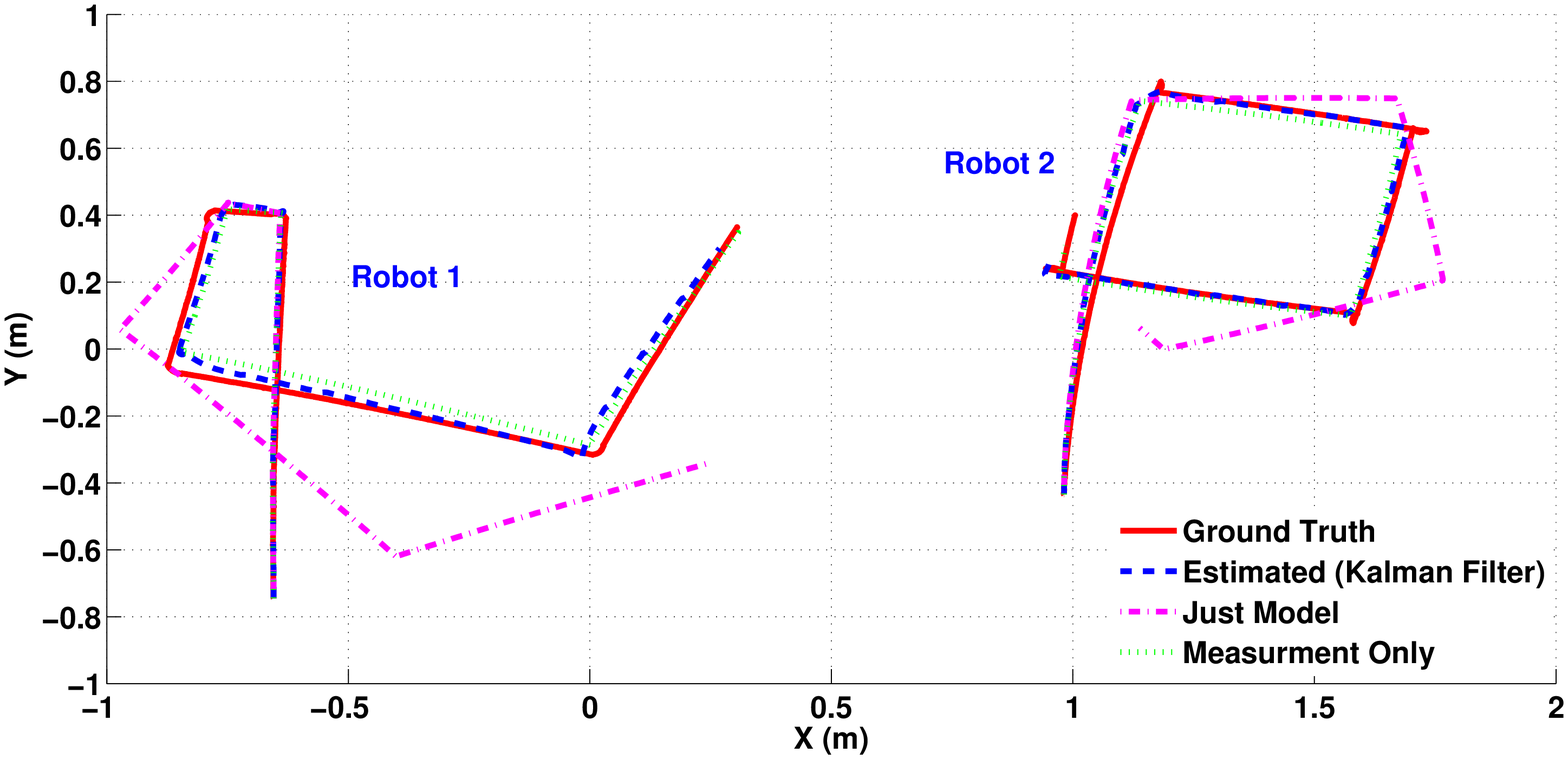}\\
  \caption{The Kalman Filtering process improves the state estimate beyond what the model and the measurements are capable of on their own.}\label{fig:kf_results}
\end{figure}

We used the information provided by the wheel encoders, optical flow sensors and the heading provided by IMU's gyroscope to estimate the position and heading. The sensor information is combined with the mathematical model of the robot to estimate the position and heading of the robot. Figure \ref{fig:kf_results} presents the result of an experiment where two robots follow a trajectory (red line) and the data from the sensors are used in the extended Kalman filter to estimate its position and heading. 
In the next two paragraphs we explain two problems we faced while implementing the designed Kalman filter and we describe the corresponding solutions used in our platform.

Timing is one of the most important factors for localization and control tasks. The delay in  communication between the robot and central computer where the Kalman filter is running, causes some variation in the time intervals within which each sensor data package is received. For example, the time intervals can vary between $50$ to $100$ $ms$ Therefore, as mentioned in section \ref{general}, the sampling time in the extended Kalman filter cannot be fixed a-priori. The state estimation procedure highly depends on the corresponding sampling time and using a fixed sampling time leads to a large error in the estimated state.One solution to this problem is to label each sensor data package with time, compute the sampling time on the central computer  and use the computed real-time sampling time in the extended Kalman filter formulation.  With these factors in mind, labeling the sensor data package with time seems to be an efficient solution to deal with variable sampling time. 
%While this approach may work in our context, it may be difficult to generalize if the timestamp cannot be effectively sent with the data.

A common problem associated with estimating the state using the velocity sensors is ``slippage''. In cases where a robot slips on the floor, the wheel encoders reflect an erroneous result. In the worst case, when the robot is stuck, the encoders keep providing ticks as if the robot is moving, which leads to a significant error in the estimated position. To solve this, we utilized the information from the optical flow sensors and designed an adaptive Kalman filter. The velocities reflected by the encoders and optical flow sensors are compared and in case there is a significant difference between them, the occurrence of slippage is inferred. Then, the noise covariance matrix in the extended Kalman filter is changed so that the filter relies more on the velocity provided by the optical flow sensor. Thus, we infer that although having multiple sensors sensing the same quantity may seem redundant, each of them, or a combination of them may be useful in different contexts.
%GENERAL LESSON 5: sensor redundancy is valuable

\subsection{Application in Real-time Control}
In a perfect scenario where there is no disturbance and model mismatch, it is possible to use some feed-forward control to drive the robot on a desired trajectory. However, in the real world, a number of issues such as model mismatch, disturbance and  error in the internal state deviates the robot from its desired path. Therefore, feedback control is vital to achieve an accurate tracking behavior. In general, the navigation problem can be devised into three categories: tracking a reference trajectory, following a reference path and point stabilization. The difference between trajectory tracking and path following is that in the former, the trajectory is defined over time while in the latter, there is no timing law assigned to it. In this section, we focus on designing a trajectory tracking controller.
%For this reason, a non-linear feedback controller is designed based on Lyapunov theory  \cite{kanayama_stable_1990}. The goal in tracking control is to reduce the error between current position and the corresponding position in the reference trajectory to zero as fast as possible considering the physical constraints of the systems such as maximum acceleration and velocity. Figure \ref{fig:trajectory} shows some \emph{experimental results} where the robot is to follow a reference trajectory (blue curve). The reference trajectory is a circle of radius $1$. The initial posture is selected to be $(0,0,0)'$. Errors $x_r-x_c$, $y_r-y_c$, and $\theta_r-\theta_c$ are shown in Figure \ref{fig:error}.

\begin{figure}[!t]
  \centering
  \includegraphics[width=0.5\textwidth]{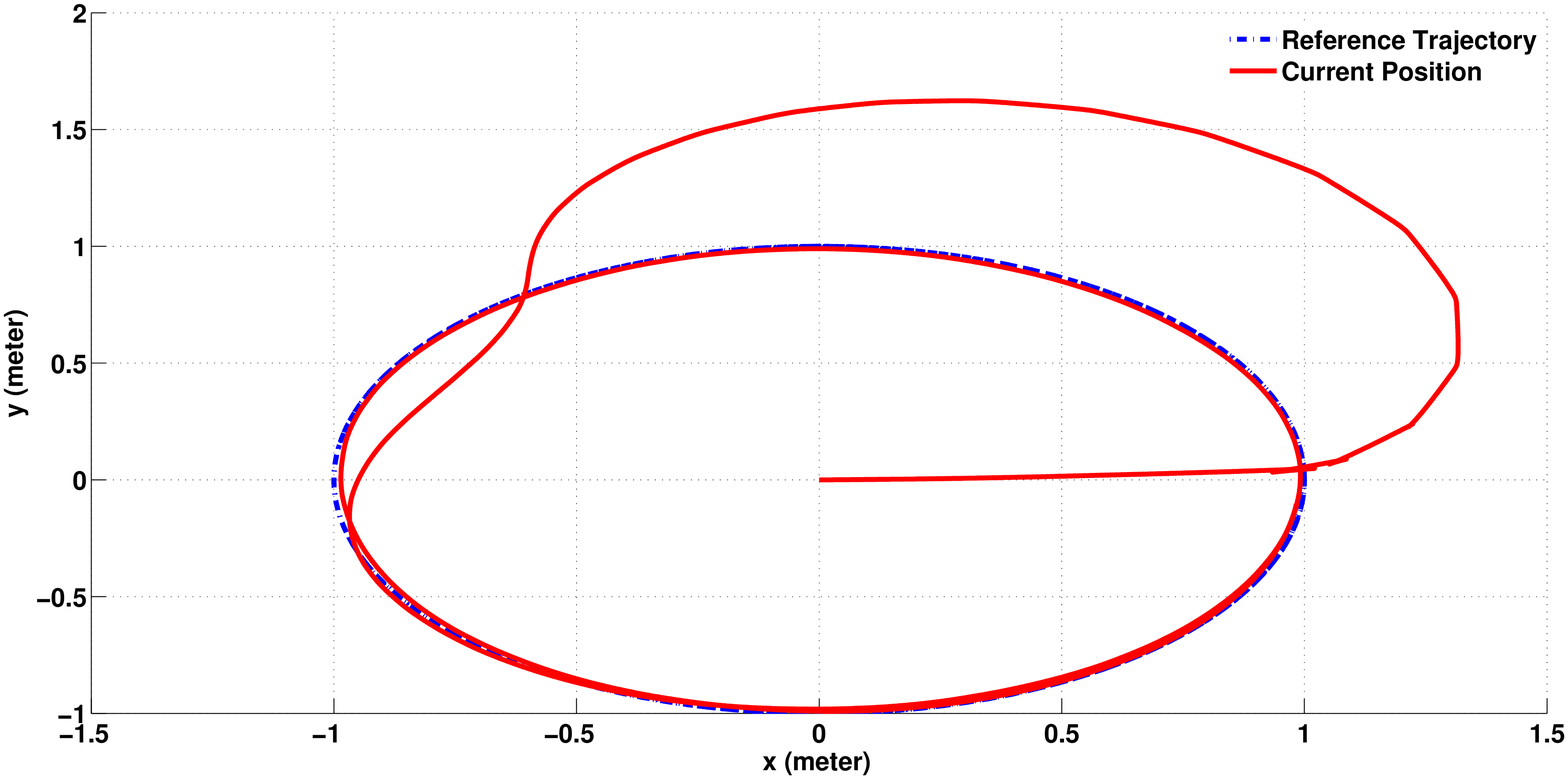}\\
  \caption{The trajectory of the robot}\label{fig:trajectory}
\end{figure}

\begin{figure}[!t]
  \centering
  \includegraphics[width=0.5\textwidth]{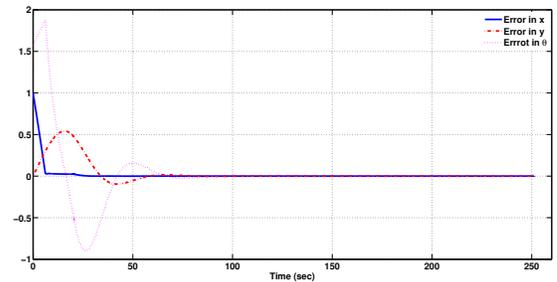}\\
  \caption{The errors in $x$, $y$ and $\theta$}\label{fig:error}
\end{figure}

The kinematic model of our platform is given by:
\begin{align}\nonumber
& \dot{x_c}=\left(\frac{v_1+v_2}{2}\right)\cos\theta_c\\\nonumber
& \dot{y_c}=\left(\frac{v_1+v_2}{2}\right)\sin\theta_c\\
& \dot{\theta_c}=\frac{1}{l}(v_1-v_2)\label{eq:system description}
\end{align}
where $v_1$ and $v_2$ are velocities of right and left wheels, $\theta_c$ is the heading (counter clockwise) and $l$ is the distance between two wheels. As stated earlier, one of the reasons for adopting a differential drive configuration for the EvoBot is the simplicity of the kinematic model (\ref{eq:system description}). Hereafter, we use two postures, namely, the ``reference posture" $p_r=(x_r,y_r,\theta_r)'$ and the ``current posture" $p_c=(x_c,y_c,\theta_c)'$. The error posture is defined as the difference between reference posture and current posture in a rotated coordinate where $(x_c,y_c)$ is the origin and the new $X$ axis is the direction of $\theta_c$.
\begin{align}
p_e=\left(\begin{array}{c}
x_e\\
y_e\\
\theta_e\\
\end{array}\right)=\left(\begin{array}{ccc}
\cos\theta_c & \sin\theta_c  & 0\\
-\sin\theta_c  & \cos\theta_c  & 0\\
0  &  0 &  1\\
\end{array}\right)(p_r-p_c).
\end{align}
The goal in tracking control is to reduce error to zero as fast as possible considering physical constraints such as maximum velocity and acceleration of the physical system. The input to the system is the reference posture $p_r$ and reference velocities $(v_r,w_r)'$ while the output is the current posture $p_r$. A controller is designed using Lyapunov theory \cite{kanayama_stable_1990} to converge the error posture to zero. It is not difficult to verify that by choosing
% \begin{align}\label{eq:the controller}
% \left(\begin{array}{c}
% v_1\\
% v_2\\
% \end{array}\right)=\left(\begin{array}{c}
% v_r\cos\theta_e+k_xx_e+\frac{l}{2}\left(w_r+v_r(k_yy_e+k_\theta\sin\theta)\right)\\
% v_r\cos\theta_e+k_xx_e-\frac{l}{2}\left(w_r+v_r(k_yy_e+k_\theta\sin\theta)\right)\\
% \end{array}\right)
% \end{align}
\begin{equation}\label{eq:the controller}
\begin{cases}
v_1=v_r\cos\theta_e+k_xx_e+\frac{l}{2}\left(w_r+v_r(k_yy_e+k_\theta\sin\theta_e)\right)\\
v_2=v_r\cos\theta_e+k_xx_e-\frac{l}{2}\left(w_r+v_r(k_yy_e+k_\theta\sin\theta_e)\right)\\
\end{cases}
\end{equation}
the resulting closed-loop system is asymptotically stable for any combination of parameters $k_x>0,\,k_y>0\text{ and } k_\theta>0$.\footnote[1]{Choosing $V=\frac{1}{2}(x_e^2+y_e^2)+(1-\cos\theta_e)$ and taking its derivative confirms that $\dot{V}\leq0$. Hence, $V$ is indeed a Lyapunov function for the system (\ref{eq:system description}).} The tuning parameters highly affect the performance of the closed loop system in terms of convergence time and the level of control input applied to the system. Hence, we chose parameters $k_x,\,k_y\text{ and}\,k_\theta$ based on the physical constraints of our platform e.g. maximum velocity and acceleration. 
Extensive simulation is performed to verify the controller (\ref{eq:the controller}). Figure \ref{fig:trajectory} shows some \emph{experimental results} where the robot follows a reference trajectory (blue curve). The reference trajectory is a circle of radius $1m$. The initial posture is selected to be $(0,0,0)'$. Errors $x_r-x_c$, $y_r-y_c$, and $\theta_r-\theta_c$ are shown in Figure \ref{fig:error}.  We remark that reference linear and angular velocities $v_r$ and $w_r$ are difficult to obtain for arbitrary trajectories. We solved this problem using some numerical methods where the point-wise linear and angular  velocities are computed numerically. 

One of the important factors in implementing the real time controller is time synchronization. The controller needs to keep track of time in order to generate the correct reference point. This issue is critical especially when a number of robots need to be  controlled. In this scenario, an independent counter can be assigned to each robot to keep track of its time and generate the correct reference point at each time step.
% \begin{itemize}
% \item
% One of the most important problems in the trajectory controller is producing the point wise reference linear and angular velocities. We used some numerical methods to get the accurate linear and angular velocities from the position vector $(x,y)$.
% \item
% migrating the controller from continuous time to discrete time needs a careful consideration. The sampling time in our setup is not fixed due to the variable delay in the communication module. Hence, all the data packages need to be time labeled and the sampling time for each data package should be computed online. The applied control command highly depends on the sampling time consequently, if we use a fixed sampling time it leads to a large error in the controller.

% Intermediate conclusion: In applications where the sampling time is not fixed, we suggest labeling the received data packages with time so that the sampling time can be computed online for each data package.

% \item
%\end{itemize}
\subsection{Application in Swarm Robotics}
The overarching application of the EvoBots platform is to enable low-cost swarm robotics research. The study of emergent collective behaviour arising from interactions between a large number of agents and their environment is an area which is gaining increasing importance recently. The robots used for these purposes are usually simple and large in number, with communication capabilities built into them. Although it is ideal to have peer-to-peer communication between the agents, the same effect can be simulated through a star-shaped network, where all the agents share information with a central machine.
%\textbf{Consensus Algorithms:} 

\begin{figure}[!t]
  \centering
  \includegraphics[width=0.99\columnwidth]{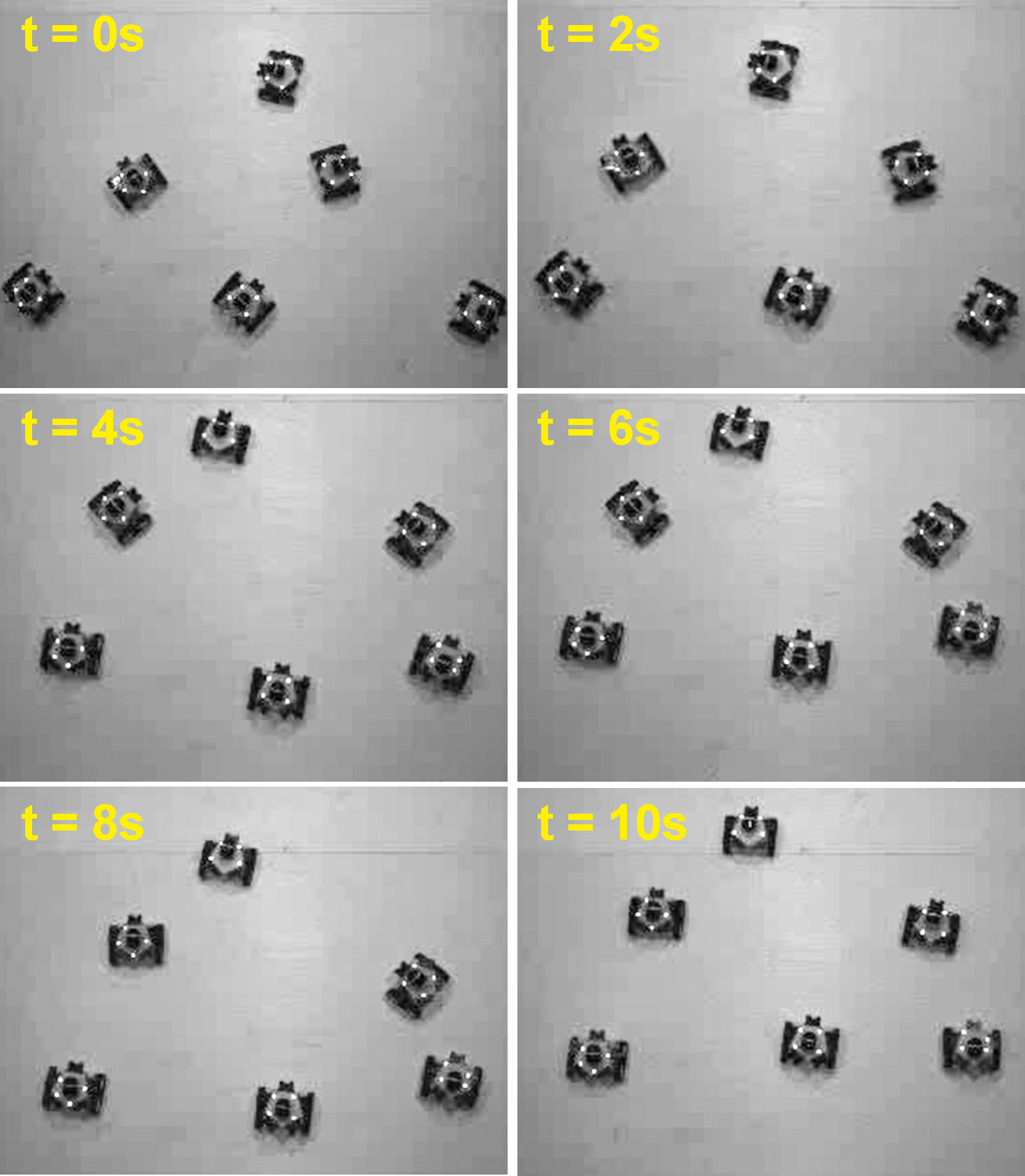}\\
  \caption{Overhead view of the robots at different times during the heading consensus. The robots are initially unaligned, but arrive at a consensus on heading at t=10s}\label{fig:theta_consensus}
\end{figure}

A typical swarm robotics application is to have the swarm arrive at a consensus about some common quantity, and to use this as a basis to collectively make decisions about the next actions of each member \cite{consensus}. 
Figure \ref{fig:theta_consensus} shows the configuration of a set of 6 robots at different times. The heading/orientation of each robot depends on those of the other robots and is governed by the update equation:\\
\begin{equation}
\theta_{i}\leftarrow\theta_{i}+K(\theta_{m}-\theta_{i})\\
\end{equation}
where \\
\begin{equation}
\theta_{m}=\sum\theta_{i}/N\\
\end{equation}
 Here, 
 $\theta_{i}$ is the heading of an individual agent, $K$ is a constant and $\theta_{m}$ is the mean heading of all $N$ agents. 
These updates are performed on each agent of the swarm till the heading converges to the same value.

As seen from Figure \ref{fig:theta_consensus}, the robots eventually orient themselves in a direction that was determined using the consensus algorithm.

%\textit{Communication rate:}
Swarm algorithms rely on the current information regarding other members. So, delays in communication can cause swarming algorithms to diverge instead of converging. As mentioned in section \ref{general}, performing routines to maintain the integrity of the exchanged data is critical. The communication rate should ideally be independent of the number of agents involved. Usually, in communication systems, it is common practice to make use of communication data buffers. For swarm applications, one should ensure the latest data is picked out either by matching similarity in time-stamps or by refreshing the buffer before each agent is queried. \\

\subsection{Application in Artificial Intelligence}

One of the long-term goals of robotics is to develop systems that are capable of adapting to unknown and unstructured environments autonomously. The general set of tools designed to enable this objective falls under the fields of machine learning and artificial intelligence (AI). This includes approaches such as neural networks \cite{Haykin:1998:NNC:521706}, support vector machines \cite{Cristianini:1999:ISV:345662}, evolutionary robotics \cite{Nolfi:2000:ERB:557168}, reinforcement learning \cite{Sutton:1998:IRL:551283}, probabilistic methods \cite{Thrun:2005:PR:1121596}, etc., which are used for a range of tasks such as object recognition, clustering, path planning, optimal control, localization and mapping etc. Although most  algorithms can be tested out in a simulated environment, the uncertainties and nuances associated with a real environment limit the utility of simulated solutions, especially in fields such as evolutionary robotics \cite{chiel1997brain,Cliff01061993,10.1371/journal.pbio.1000292}. A more thorough validation necessitates implementation of these algorithms on real-world physical systems.\\
To illustrate some of the primary issues and requirements of a platform for use in AI, a path planning task is demonstrated using the standard A-star algorithm \cite{4082128}.
The scenario involves planning a path from the starting position of a robot to a target position using a previously constructed infra-red sensor based map of the environment as shown in Figure \ref{fig:path_planning}. Most of the arena shown in Figure \ref{fig:path_planning} has been mapped; the mapped areas are shown as black patches along the boundaries of the arena. The planned path (shown in blue) is generated by the A-star algorithm, using cost and heuristic functions. The heuristic function is proportional to the euclidean distance from the target position. 
Some of the important issues while implementing this algorithm for path planning and for AI applications in general, are listed below:

\begin{figure}[!t]
  \centering
  \includegraphics[width=0.99\columnwidth]{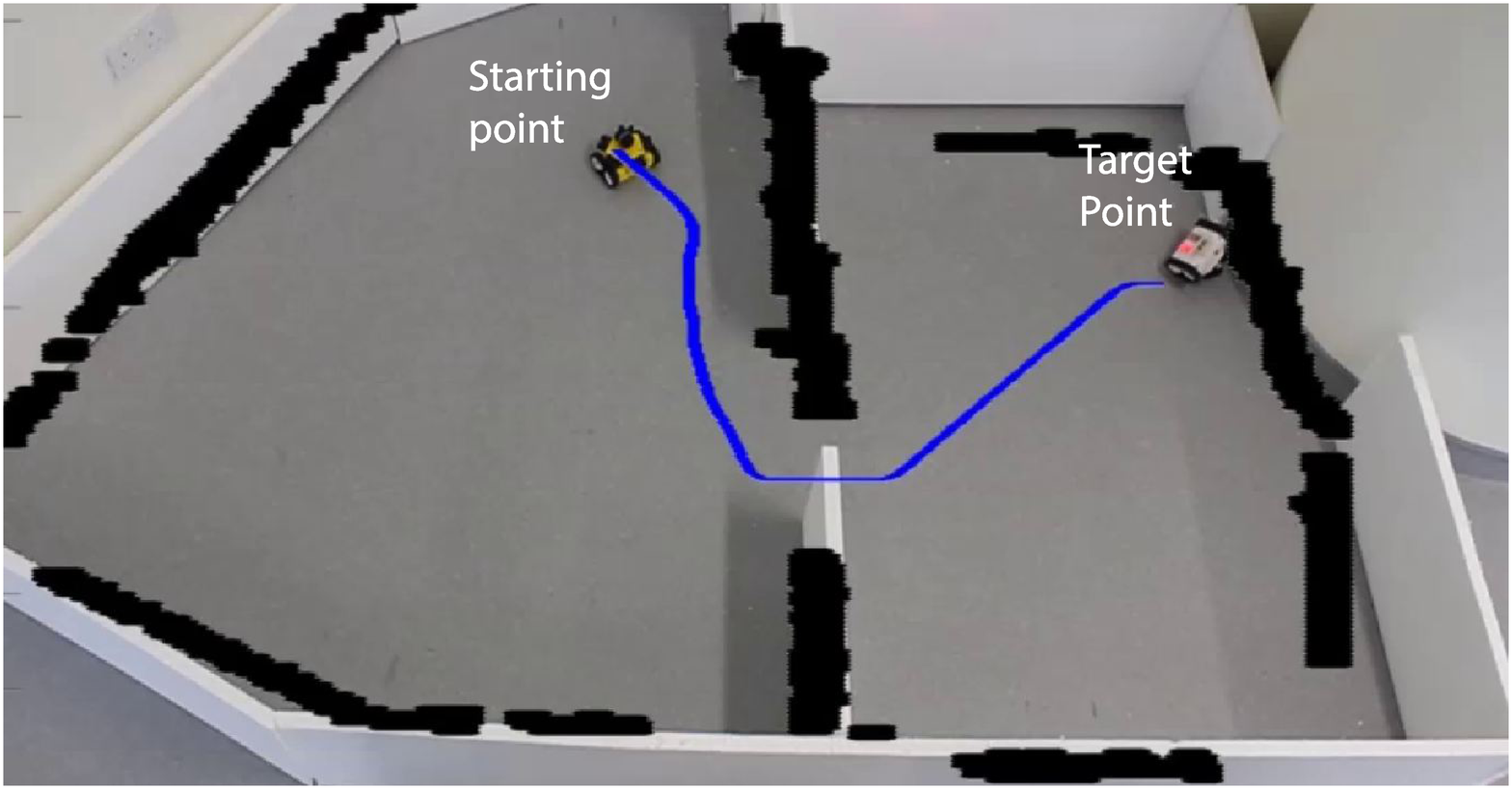}\\
  \caption{Planned path of the robot shown in blue}\label{fig:path_planning}
\end{figure}

\textit{Computing Resources:}
The A-star algorithm could have also been implemented locally on the robots. However, this would have implied storing a copy of the IR map and performing the A-star algorithm locally using this map. This may not have been feasible due to the limited on-board memory and computational complexity of the algorithm. \\
  Also, for path planning, the map of the environment gets larger as the space explored gets larger. Accordingly, the A-star path planning could potentially become more computationally intensive. So setting a resolution parameter (or deciding to have one) is recommended before starting the mapping.\\
   In general, sufficient computing resources will be needed to carry out AI algorithms. Having a thin client system that offloads the computational effort to a central server helps deal with the high computational costs imposed due to high dimensionality in certain problems.

\textit{Dealing with the real world:}
Readings from the robot's sensors involve an inherent component of noise, even after they have been calibrated. In addition, it may be common for one to consider an agent as a point moving through a space during the simulation stage. In reality, the agent cannot be considered to be a point in space.These aspects of reality could potentially hinder one from achieving the desired results. Taking the physical dimensions of its body into consideration is one way to counter this.\\
  In the above mentioned case of path planning, noise points  were first removed using median filters from the raw map. Also, in order to introduce a safety margin, (and to compensate for the loss of information introduced due to blind spots as well as due to the resolution parameter) the detected obstacles were inflated by exaggerating their size. The safety margin could also be expanded to account for the dimensions of the robot body, thus ensuring a smoother transition from simulation to reality, without having to explicitly encode the details of the body geometry during simulation, thereby saving considerable development time. It should be noted that excessive expansion of obstacles could lead to failure of A-star to find a valid path to the goal. \\
  Noise, and mismatches between virtual and real environments is inevitable. Noise removal using appropriate filters, and compensating for mismatches with the real world by erring on the side of caution may be considered to be useful measures to ensure a smooth transition from virtual to real environments.

\section{Conclusion}
This paper discussed solution strategies to address a set of common problems which may be encountered while adapting a low-cost robotics platform such as the EvoBot for research. Some insights into the factors to be considered while choosing the sensing, control and communication capabilities of low-cost platforms were discussed. The eventual success of our approach was described in the context of four diverse yet common robotics tasks - state estimation, real-time control, a basic swarm consensus application and a path planning task that were carried out using the EvoBots platform. \\Some of the potential issues faced by end users who are not familiar with real-world implementation of algorithms are also discussed and potential work-arounds for a smooth transition from simulation to reality were suggested. The points discussed in this paper aim to serve as a guide to groups who intend to adapt similar low-cost platforms in the future, as well as to researchers working on theoretical aspects of robotics, who intend to validate their algorithms through real-world implementation.  

\bibliographystyle{plain}
\bibliography{ref}

\end{document}